\documentclass[runningheads]{llncs}
\usepackage[T1]{fontenc}

\usepackage{graphicx}
\usepackage{svg}

\usepackage{amsmath}
\usepackage{amsfonts}
\usepackage{amssymb}
\usepackage{upgreek} 
\usepackage{xcolor}
\usepackage{hyperref}
\usepackage[capitalise,nameinlink]{cleveref} 
\crefname{figure}{Fig.}{Figs.} 
\Crefname{figure}{Figure}{Figures}
\crefname{table}{Table}{Tables}
\Crefname{table}{Table}{Tables}
\usepackage{subcaption}
\usepackage{booktabs}
\begin{document}
\title{Temporal Periodic Image Registration with Implicit Neural Representations}

%
\author{Mathias Lowes \inst{1}\orcidID{0009-0003-0002-1645} 
\and
Kristine Aavild Sørensen \inst{1,2} \orcidID{0000-0003-1434-697X}
\and
 Maxime Sermesant\inst{3} \orcidID{0000-0002-6256-8350}
 \and
 Rasmus R. Paulsen\inst{1}\orcidID{0000-0003-0647-3215}
 }
\authorrunning{M. Lowes et al.}
%
\institute{Department of Applied Mathematics and Computer Science, Technical University of Denmark, Kgs. Lyngby, Denmark \\ \email{mmilo@dtu.dk}\and
Novo Nordisk A/S, Copenhagen, Denmark \and
Inria Université Côte d’Azur, Nice, France
}

\maketitle              

\begin{abstract}
Implicit Neural Representations (INRs) have recently gained popularity for their ability to model functions using simple and lightweight networks. These methods have also proven effective in deformable image registration.
In this work, we extend INRs for deformable image registration to the domain of temporal image registration, focusing on periodic temporal image sequences.
Our approach, Temporal-IDIR, optimizes a single INR to model deformations across all frames in a temporal image sequence simultaneously, allowing for self-regularization through its own deformation predictions. To achieve this, we introduce a temporal consistency loss that penalizes discrepancies between direct source-to-target transformations and those traversing intermediate frames.
We evaluate our framework on the DIR-LAB dataset, using the target registration error (TRE) between annotated and moved landmarks as the metric. Here, we achieve a TRE of 1.03 mm, outperforming other INR-based registration methods.
Additionally, our framework supports smooth interpolation between time frames by estimating deformations between the given input frames.\footnote{Code is publicly available at \url{https://github.com/MMLowes/Temporal_INR}.}

\end{abstract}
%
        

\section{Introduction}
Temporal image registration is a fundamental process in image analysis that involves aligning two or more images taken at different times.
In medical imaging, temporal registration is useful for analyzing short-term sequences, such as a heart beat, a respiratory cycle, or bowel movements, as well as long-term sequences to investigate disease progression through months or years. 
Registering the images across the temporal sequence offers the possibility to propagate landmarks and thereby follow the movement of anatomical structures over time. 

\begin{figure}[t]
    \centering
    \includegraphics[width=\linewidth]{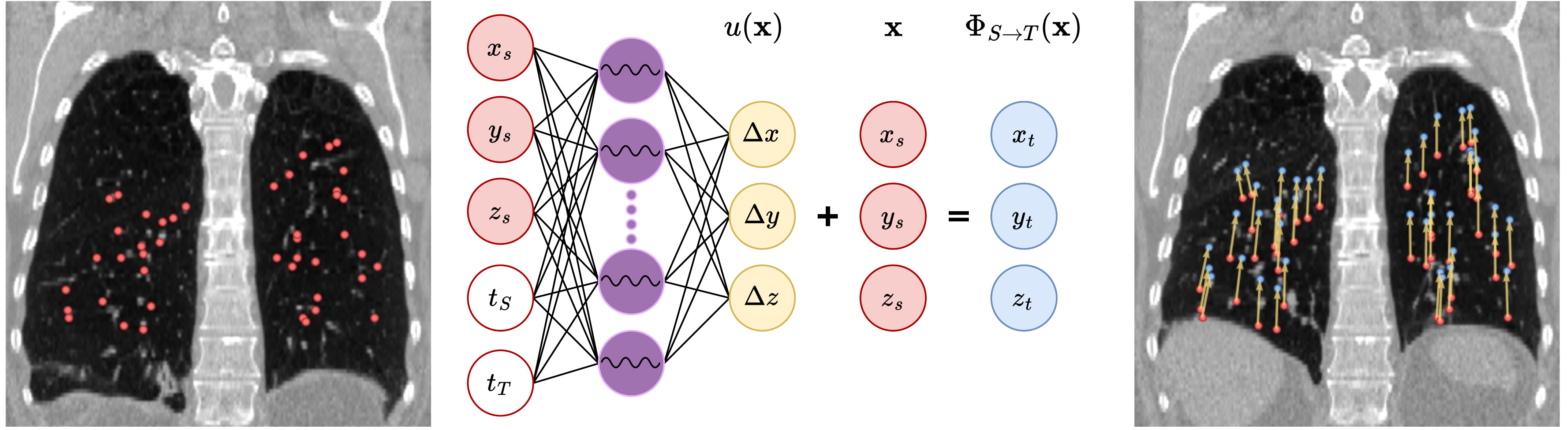}
    \caption{Registration of multiple landmarks from source frame (left) to target frame (right). For each source point $\mathbf{x}$ the deformation $u(\mathbf{x})$ is calculated to map the point to the target domain $\mathrm{\Phi}_{S\to T} (\mathbf{x}) $ }
    \label{fig:overview}
\end{figure}

Standard deformable image registration (DIR) is performed pairwise by registering a source image $I_S$ to a target image $I_T$.
This can be achieved by applying a deformable vector field $u(\textbf{x})$ to map each position in the source image $\textbf{x}$ to align it with the target image (see Figure \ref{fig:overview}). 
Registration methods can be divided into two categories; \textit{optimization-based} and \textit{learning-based} methods. 
Traditional methods using B-splines to represent the deformation vector field \cite{Klein2010elastix,Rueckert1999FFD,Wu2016Bspline} are examples of optimization-based methods, where the parameters of the splines are optimized individually for each registration. 
With convolutional neural networks, a group of learning-based registration methods appeared \cite{balakrishnan2019voxel,he2021rrn,Heinrich2022voxelPlusPlus,Sokooti2017cnn,deVos2019cnnreg,Yang2017Patchbased}. 
This approach offers the possibility of learning the registration on one data set and generalize to new unseen data, however using 3D CNNs on large 3D image volumes requires large amounts of memory. 

Implicit Neural Representations (INRs) can be used as an optimization-based registration method by parameterizing the deformation field as a neural network \cite{wolterink2021implicit}. 
In the simplest form, the source point coordinate $\textbf{x}$ is passed to a simple multi layer perceptron (MLP) that estimates the deformation $u(\textbf{x})$. 
This approach significantly lowers the memory requirements by operating on continuous points, while exploiting the neural networks ability to approximate any signal or function.
Several works have expanded on the use of INRs by enforcing an inverse consistency between source-to-target and target-to-source registration \cite{Byra2023INRnature,vanharten2024cycle,tian2023gradicon} or applying geometric constraints \cite{harten2023deformable}.

The described methods are developed for pair-wise image registration. 
When a temporal sequence consists of more than two images, one can utilize pair-wise registration to either register the first image to all subsequent time frames (\textit{direct registration}) or sequentially register each time frame to the next (\textit{sequential registration}). 
Assuming smooth changes over the temporal sequence, the sequential registration will only require small deformations between consecutive time steps, whereas the direct registration may require increasingly larger deformations when the time difference is large. 
The sequential approach, however, builds up error through the time steps, causing problems in long sequences \cite{lowes2025inrsdf}. 

 

In this paper, we propose a novel method for temporal registration using INRs that combines direct and sequential registration, by simultaneously optimizing registration between all possible image pairs in a temporal image sequence.
Our main contributions include extending INR-based registration with the temporal dimension, enabling tracking of landmark movement during a whole cycle of data, contrary to only the extreme phases. Furthermore, our method tracks landmarks between the input frames of the data, giving continuous landmark movement throughout the data cycle.
We demonstrate the capabilities of the method on the open source DIR-LAB lung CT dataset~\cite{Castillo2009DIRLAB}, and show that the proposed method outperforms other INR-based registration methods. 

\section{Methods}
\textbf{Image Registration.}
Pairwise deformable image registration aims to find a mapping from a source image $I_S : \Omega \subset \mathbb{R}^3 \rightarrow \mathbb{R}$ to a target image $I_T : \Omega \subset \mathbb{R}^3 \rightarrow \mathbb{R}$. 
The mapping $\mathrm{\Phi}_{S\to T}: \mathbb{R}^3 \rightarrow \mathbb{R}^3$ is expressed as $\mathrm{\Phi}_{S\to T}(\mathbf{x})=u(\mathbf{x})+\mathbf{x}$, where $u(\mathbf{x})$ represents the displacement field at the point $\mathbf{x}$. The goal of the pairwise registration is to satisfy
\begin{equation}
    (I_S \circ \mathrm{\Phi}_{S\to T})(\mathbf{x})=I_T(\mathbf{x}) \text{ for all } \mathbf{x} \in \Omega.
\end{equation}

In scenarios involving temporal image sequences, temporal registration is employed to map images over the given sequence of time points. This can be performed using two approaches:
\begin{itemize}
    \item \textbf{Direct registration}, aligning a single source frame to all other frames in the sequence e.g. $f_0 \to f_1$, $f_0 \to f_2$, etc.
    \item \textbf{Sequential registration}, aligning consecutive frames to each other, such as $f_0 \to f_1$, $f_1 \to f_2$, etc.
\end{itemize}
To take advantage of both strategies, we propose a unified optimization framework that simultaneously optimizes the registrations between all source-target pairs.
In this way, the temporal registration, i.e. the mapping between all source and target frames, $\mathrm{\Phi}$, is found by optimizing

\begin{equation} \label{eq:reg}
    \mathrm{\Phi} ^*=\underset{\mathrm{\Phi}}{\operatorname*{arg\,min}}\Biggl\{
    \sum_{S,T}
    \mathcal{L}_{\text{sim}}(I_S \circ \mathrm{\Phi}_{S\rightarrow T}, I_T)
    \Biggr\},
\end{equation}
where $\mathcal{L}_{\text{sim}}$ is a similarity loss, chosen as the normalized cross-correlation loss. 

\textbf{Temporal Consistency Loss.}
To ensure point correspondence over all temporal mappings, we introduce a temporal consistency loss, $\mathcal{L}_{\text{temporal}}(\mathrm{\Phi}_{S \to T})$. This loss enforces that a point moved directly from $S$ to $T$ aligns with the same point moved stepwise through intermediate frames from $S$ to $T$. The loss is defined as
\begin{equation}\label{eq:temporal_loss}
    \begin{aligned}
        \mathcal{L}_{\text{temporal}}(\mathrm{\Phi}_{S\to T}) 
        = &\left\Vert \mathrm{\Phi}_{S\rightarrow T}(\mathbf{x})-\mathrm{\Phi}_{S\to T}^+ (\mathbf{x}) \right\Vert_2 \\
        + &\left\Vert \mathrm{\Phi}_{S\rightarrow T}(\mathbf{x})-\mathrm{\Phi}_{S\to T}^- (\mathbf{x}) \right\Vert_2 \\
        + &\left\Vert \mathrm{\Phi}_{S\to T}^+ (\mathbf{x})-\mathrm{\Phi}_{S\to T}^- (\mathbf{x}) \right\Vert_2 ,
    \end{aligned}
\end{equation}
where $\mathrm{\Phi}_{S\to T}^+$ and $\mathrm{\Phi}_{S\to T}^-$ is the sequential movement of point $\mathbf{x}$ from $S$ to $T$ in the forward and backwards directions, defined by
\begin{equation}\label{eq:sequential_operator}
    \begin{aligned}
        \mathrm{\Phi}_{S\to T}^+ (\mathbf{x}) &= \mathrm{\Phi}_{T-1 \to T} \circ \ldots \circ \mathrm{\Phi}_{S+1 \to S+2} \circ  \mathrm{\Phi}_{S \to S+1}(\mathbf{x}), \\
        \mathrm{\Phi}_{S\to T}^- (\mathbf{x}) &= \mathrm{\Phi}_{T+1 \to T} \circ \ldots \circ \mathrm{\Phi}_{S-1 \to S-2} \circ  \mathrm{\Phi}_{S \to S-1}(\mathbf{x}) .
    \end{aligned}
\end{equation}
Given the periodic nature of the data sequence, modular indexing is used over the sequence length $N$, to ensure that $\mathrm{\Phi}_{N-1\to N}$ is equivalent to $\mathrm{\Phi}_{N-1\to 0}$.

For regularization, we adopt the symmetric Jacobian regularization loss from van Harten et al.~\cite{vanharten2024cycle} penalizing the Jacobian matrix's deviation from 1.

\begin{equation}
    \mathcal{L}_{\mathrm{reg}}(\mathrm{\Phi}_{S\to T}) = \min \left\{ \frac{\left(\operatorname{det} \nabla \mathrm{\Phi}_{S\to T}-1\right)^2}{| \operatorname{det} \nabla \mathrm{\Phi}_{S\to T} |}, \tau\right\},
\end{equation}
where  $\nabla \mathrm{\Phi}_{S \to T}$ is the Jacobian matrix of $\mathrm{\Phi}_{S\to T}$ and $\tau$ is a hyperparameter which clips the regularization. As the authors suggest, we use $\tau=10$.

With the inclusion of the temporal loss and the regularization term, the registration over a periodic temporal image sequence is found as
\begin{multline}\label{eq:opt_final}
    \mathrm{\Phi} ^*=\underset{\mathrm{\Phi}}{\operatorname*{arg\,min}}\Biggl\{
    \sum_{S,T} \bigl[
    \mathcal{L}_{\text{sim}}(I_S \circ \mathrm{\Phi}_{S\rightarrow T}, I_T)
    + \alpha \mathcal{L}_{\text{sim}}(S_S \circ \mathrm{\Phi}_{S\rightarrow T},S _T) \\
    +\beta \mathcal{L}_{\text{temporal}} (\mathrm{\Phi}_{S\rightarrow T})
    +\lambda \mathcal{L}_{\text{reg}}(\mathrm{\Phi}_{S\rightarrow T}) 
    \bigr]
    \Biggr\}.
\end{multline}
With $\alpha$, $\beta$ and $\lambda$ as hyperparameters for the weight of the regularization terms. The signed distance fields (SDFs) of the anatomy of interest (for DIR-LAB the lungs), are included in the optimization and denoted $S_S$ and $S_T$ for the  source and target SDFs, respectively.

\textbf{Implicit Neural Representations.}
An INR consists of a simple MLP network, trained to approximate the deformation fields between all frames in the data sequence.
The INR takes a continuous coordinate $\mathbf{x}$ and source and target times as input, while returning the mapping $\mathbf{x}$ from source to target, see~\cref{fig:overview}.

To enable the neural network to capture both low and high frequency signals we use the \texttt{SIREN} architecture~\cite{sitzmann2019siren}, an MLP with sinusoidal activation functions. Additionally, the physical input coordinates are encoded with a harmonic encoding, projecting the coordinates into a higher-dimensional space, while the input times are mapped to the unit circle.
The encoded physical coordinates, $\gamma(\mathbf{x})$, and the encoded time coordinates, $\psi(t)$, are calculated as
\begin{gather}    \label{eq:temporal_input}
    \gamma(\mathbf{x})=\left[\mathbf{x}, \cos(2^0\mathbf{x}),\sin(2^0\mathbf{x}), \ldots, \cos(2^{L-1}\mathbf{x}),\sin(2^{L-1}\mathbf{x})\right], \\
        \psi(t)=\left[\cos\left(\frac{2\pi}{N}t\right),\sin\left(\frac{2\pi}{N}t\right)\right], 
\end{gather}
where $L$ is the number of harmonic encodings and $N$ is the cycle length.

\textbf{Temporal Inference.} 
The registrations are evaluated based on the error between landmark in expiratory phase and landmark moved from inspiratory phase. We employ four methods for mapping from the inspiratory phase $t_\text{insp}$ to the expiratory phase $t_\text{exp}$.
We define $\mathbf{x}_\text{insp}$ as the landmark position in the inspiratory phase and $\tilde{\mathbf{x}}_\text{exp}$ as the estimated moved landmark position after applying the transformation.
However, the methods can be used to map from any source time to any target time.

\begin{enumerate}
    \item 
The direct method calculates the transformation from $t_\text{insp}$ to $t_\text{exp}$ as
\begin{equation}
    \tilde{\mathbf{x}}_\text{exp} = \mathrm{\Phi}_{t_\text{insp} \to t_\text{exp}}(\mathbf{x}_\text{insp}) .
\end{equation}
\item 
To move the inspiratory landmark sequentially we use the definition from \cref{eq:sequential_operator}, moving the point one step at a time
\begin{equation}
    \tilde{\mathbf{x}}_\text{exp} = \mathrm{\Phi}^+_{t_\text{insp} \to t_\text{exp}}(\mathbf{x}_\text{insp}) .
\end{equation}
\item 
Combining the direct and sequential methods aims to utilize more of the information the network has learned during training. The given point is moved one step sequentially forward and averaged with the point moved directly from $t_\text{insp}$ to the given time. This method relies on the previous point, thus it has to be calculated for intermediate points $\tilde{\mathbf{x}}_i$ as well
\begin{equation}
    \tilde{\mathbf{x}}_{i} = \frac{1}{2}\left(\mathrm{\Phi}_{t_\text{insp} \to t_{i}}(\mathbf{x}_\text{insp}) +\mathrm{\Phi}_{t_{i-1} \to t_{i}}(\tilde{\mathbf{x}}_{i-1})\right).
\end{equation}
\item 
The Taylor method for inference leverages the inverse consistency between the direct mapping from source to target and vice versa~\cite{vanharten2024cycle}. 
The mapping from source to target is averaged with the inverse mapping of the target to source mapping, which is approximated with a Taylor expansion
\begin{equation}
    \tilde{\mathbf{x}}_\text{exp} = \frac{1}{2}\left(\mathrm{\Phi}_{t_\text{insp} \to t_\text{exp}}(\mathbf{x}_\text{insp}) +\mathrm{\Phi}^{-1}_{t_\text{exp} \to t_\text{insp}}(\mathbf{x}_\text{insp})\right),
\end{equation}
where $\mathrm{\Phi}^{-1}_{t_\text{exp} \to t_\text{insp}}$ is is approximated with a first-order Taylor expansion. For ease of notation we will use $\mathrm{\Phi}_{F}=\mathrm{\Phi}_{t_\text{insp} \to t_\text{exp}}$, $\mathrm{\Phi}_{B}=\mathrm{\Phi}_{t_\text{exp} \to t_\text{insp}}$ and $\mathbf{x}=\mathbf{x}_\text{insp}$, representing the forward and backward mappings and the inspiratory landmark, respectively. The inverse backward mapping is calculated as
\begin{equation} \label{eq:tayler_exp}
    \mathrm{\Phi}^{-1}_B(\mathbf{x}) =  \mathrm{\Phi}_{F}(\mathbf{x}) +\nabla^{-1}\mathrm{\Phi}_{B}(\mathrm{\Phi}_{F}(\mathbf{x}))(\mathbf{x}-\mathrm{\Phi}_{B}(\mathrm{\Phi}_{F}(\mathbf{x})),
\end{equation}
where $\nabla^{-1}\mathrm{\Phi}_{B}$ is the inverse Jacobian matrix calculated at the point $\mathrm{\Phi}_{F}(\mathbf{x})$.
\end{enumerate}

\section{Experiments and Results}
\textbf{Dataset.}
The DIR-LAB data set~\cite{Castillo2009DIRLAB} consists of 4D CT scans of the lungs of 10 participants. Each 4D CT scan is composed of 10 volumetric frames $(f_0,\ldots, f_{9})$ where the extreme phases, the inspiratory and expiratory phases, are at $f_0$ and $f_5$, respectively. 
Registration from inspiratory phase to expiratory phase in DIR-LAB is commonly used to validate image registration for medical imaging~\cite{balakrishnan2019voxel,Hansen2021graphregnet,vanharten2024cycle,he2021rrn,tian2023gradicon,Tian2023NePhi,Vishnevskiy2017ptv,wolterink2021implicit}. In the two extreme phases of each 4D CT, 300 expert annotated landmarks are available to measure registration performance.
As common practice, we only calculate the deformation field within the lungs, with a mask obtained from Hofmanninger et al.~\cite{hofmanninger2020lung}.
The DIR-LAB data set also contains 75 landmarks per patient that have been propagated from frame 0 and frame 5 to the intermediate frames, which is used to evaluate the outside the extreme phases. However, as these are not expert annotated, the results from these are not as reliable.

\textbf{Experimental Details.}
The INRs used for the cyclic registration are parameterized with 5 hidden layers each with 512 neurons. They are trained for 10~000 epochs using the Adam optimizer\cite{Kingma2014AdamAM} with a learning rate of $10^{-4}$, decaying every 1000 epochs by a factor of 0.5.
An epoch consists of 15 000 randomly sampled points within the union of lung masks for all time frames. A source and a target time are randomly sampled between 0 and 9, to accommodate each point, here the extreme phases are sampled more frequently. 

For our optimal model, we use the hyperparameters $\alpha=0.05$, $\beta=0.01$, and $\lambda=0.005$ for \cref{eq:opt_final}. We test multiple values of $\beta$, while the other parameters are taken from other INR registration frameworks.
The methods are evaluated using the target registration error (TRE) between annotated landmarks in the expiratory phase and landmarks moved from inspiratory to expiratory.
Registration over a whole cycle takes 12.5 min on an NVIDIA Geforce RTX 4090.

\begin{table}[t]
\centering
\caption{Comparison of mean target registration error (TRE) with other registration methods for landmark registration inspiration to expiration. The results for the other methods are adopted from the literature.}
\label{tab:results:comparison}
\small
\begin{tabular}{lc}
\toprule
Method & TRE [mm] $\downarrow$ \\
\midrule
VoxelMorph+~\cite{balakrishnan2019voxel}    & 2.89 \\
\texttt{GradICON}~\cite{tian2023gradicon}   & 1.31 \\
GraphRegNet~\cite{Hansen2021graphregnet}    & 1.39 \\
PTVReg~\cite{Vishnevskiy2017ptv}    & 0.96 \\
RRN~\cite{he2021rrn}                & 0.83 \\
\midrule
IDIR~\cite{wolterink2021implicit}   & 1.07 \\
Cycle-IDIR~\cite{vanharten2024cycle}   & 1.04 \\
\texttt{NePhi}~\cite{Tian2023NePhi}    & 1.46 \\
\midrule
Temporal-IDIR (ours)                & 1.03\\
\bottomrule
\end{tabular}
\end{table}

\begin{table}[t]
\centering
\caption{Evaluation of $\beta$-values across the four inference methods in mean registration errors in mm for the landmark registration.}
\label{tab:results:inr}
\setlength{\tabcolsep}{8pt}
\begin{tabular}{@{}lcccc@{}}
\toprule
Setting     & Direct & Sequential     & Combined      & Taylor        \\ \midrule
$\beta=0.0$    & 1.09   & 1.52           & 1.09          & 1.05          \\
$\beta=0.01$ & \textbf{1.08}   & 1.38           & 1.08          & \underline{\textbf{1.03}} \\
$\beta=0.05$ & 1.09   & 1.21           & 1.08          & 1.04          \\
$\beta=0.1$  & \textbf{1.08}   & 1.16           & \textbf{1.06} & 1.04          \\
$\beta=0.5$  & 1.09   & \textbf{1.11} & 1.08          & 1.07          \\
$\beta=1.0$  & 1.10    &    1.12 &   1.12  &     1.10 \\
$\beta=2.0$ &  1.23    &    1.23 &     1.23 &  1.22 \\
\bottomrule
\end{tabular}
\end{table}

\begin{figure}[ht]
    \centering
    \includegraphics[width=\linewidth]{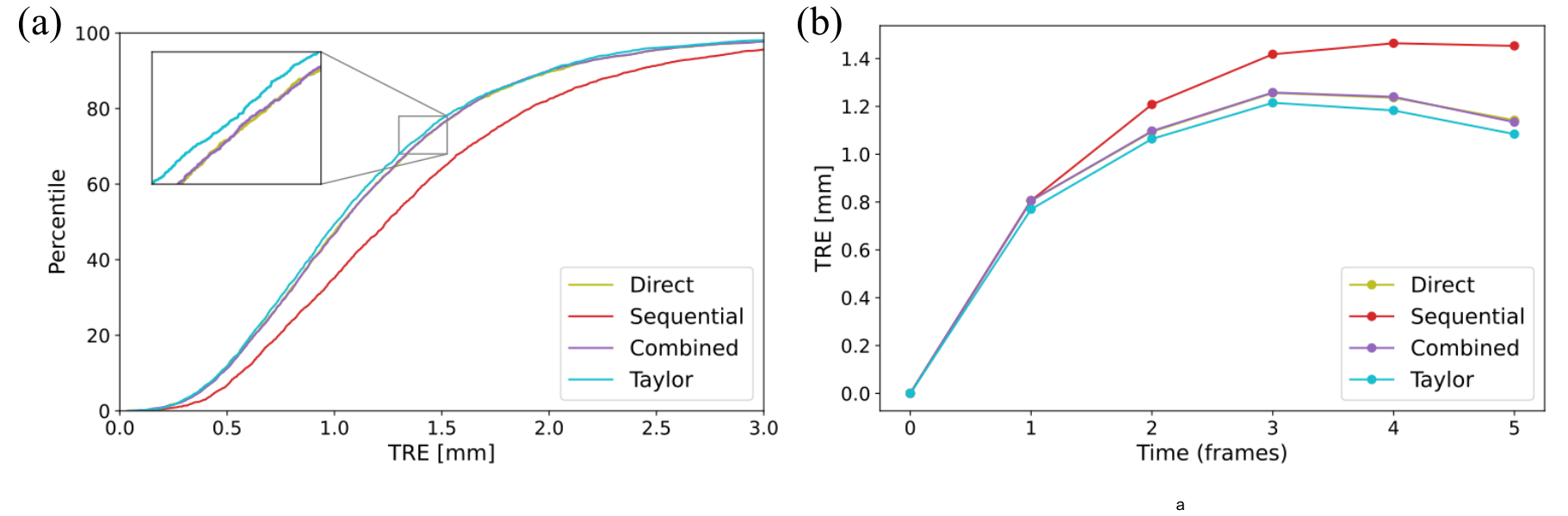}
    \caption{Comparison of interference methods for model with $\beta=0.01$. (a) cumulative distribution of TRE for all in the dataset. (b) Mean TRE over time. }
    \label{fig:method_comparison}
\end{figure}

\textbf{Registration Results.}
In \cref{tab:results:comparison} we compare our method to other state-of-the-art registration methods on the DIR-LAB dataset including several INR-based methods. The results are taken from their respective papers except VoxelMorph+~\cite{balakrishnan2019voxel} where the testing was conducted by Hansen et al.~\cite{Hansen2021graphregnet}.
Our method beats the other INR-based methods, but not the PTVReg~\cite{Vishnevskiy2017ptv} and RRN~\cite{he2021rrn} methods. 



\begin{figure}[t]
    \centering
    \includegraphics[width=0.9\linewidth]{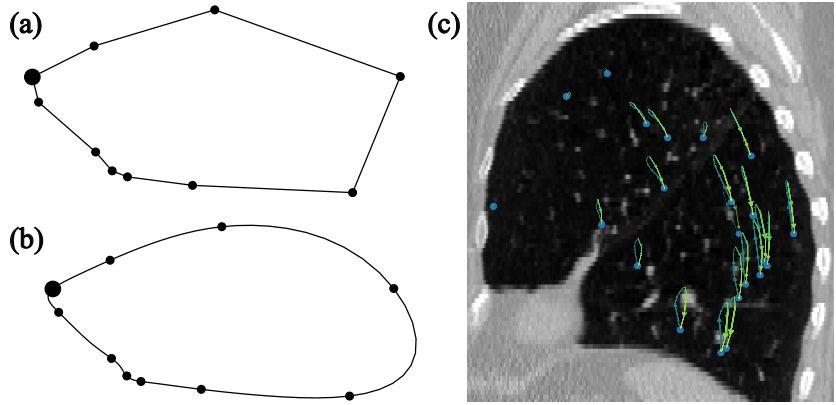}
    \caption{Landmark movement in the sagittal plane (yz). (a) and (b) illustrates the movement of a single point using 10 time steps and 100 time steps, respectively. (c) illustrates the movement of several points over a respiratory cycle.}
    \label{fig:cycle_combined}
\end{figure}

\Cref{tab:results:inr} explores both the sensitivity to changes in the $\beta$-hyperparameter, i.e. the temporal loss, and the precision of the four proposed inference methods. 
The framework is quite robust to changes in the $\beta$-hyperparameter, with lowest registration error with the Taylor method across all experiments, and lowest with $\beta=0.01$. The registration errors for this model are illustrated as a cumulative distribution for each method in in \cref{fig:method_comparison}(a) and as error over time in \cref{fig:method_comparison}(b).

The estimated landmark movement throughout the respiratory cycle is illustrated for several landmarks in \cref{fig:cycle_combined}(c).
\Cref{fig:cycle_combined}(a,b) illustrates how the network captures the movement between the training frames of the 4D CT. 


\section{Discussion and Conclusion}
Our proposed framework, Temporal-IDIR, extends INR-based registration to periodic temporal data, by including two timesteps in the input to the INR and training for registration between all frames simultaneously. This enables the framework to model the movement between any two given time points in the 4D CT scans.
With the inclusion of the temporal consistency loss, \cref{eq:temporal_loss}, in combination with the Taylor inference method, the network has a slight edge over the other INR-based registration methods on the DIR-LAB dataset. 

Although our framework does not outperform the two methods RRN~\cite{he2021rrn} and pTVreg~\cite{Vishnevskiy2017ptv} in the TRE metric, it has other advantages. INR based methods are not restricted to a grid as both these methods are, but operates on continuous points. This entails that the network size and memory load during optimization are more consistent over inputs with different sizes and resolutions.

The temporal loss $\mathcal{L}_{\text{temporal}}$ is controlled with the $\beta$-hyperparameter. Having a small value of $\beta$ generally gives the best results as seen in \cref{tab:results:inr}, where especially the Taylor method obtain a low TRE. The sequential method is not on par with the other methods for small $\beta$-values. Small errors propagate through multiple deformations, resulting in larger deviations when registering between extreme phases. The temporal loss penalizes this propagation of errors. Thus, we see an increase in the performance of the sequential method when increasing $\beta$, until it is on par with the direct method, at $\beta=1.0$. At this point, there is not more to gain from the temporal loss. 
The increased performance of the sequential method can, in some cases, make this method preferred to the other methods. Each deformation is smaller with the sequential method; this can give better regularity over the deformations, potentially increasing the quality of a moved object compared to moving it with one big deformation.


As the temporal inputs to the model are sampled from the unit circle, our framework is capable of modeling the movement of landmarks continuously, as opposed to only modeling the movement between extreme phases. 
The network is trained with 10 different temporal inputs, corresponding to the 10 frames in the 4D CT scans. However, as \cref{fig:cycle_combined}(a,b) shows, we are able to input any encoding from the unit circle as the temporal input. This enables our method to interpolate between the 10 time frames on which it was trained and to do a continuous smooth movement of landmarks. 
In general, the framework can be trained with input with any number of temporal dimensions, but inference can also be done with higher resolution than the training process.
The ability to model movement over the whole sequence of the data makes our method better suited for practical use, as clinical cases involving organ movement, e.g. the heart movement, the whole temporal movement is of interest.

In conclusion, our INR-based temporal registration method is able to capture the movement of the lungs over a respiratory cycle. The registration quality is better than other INR-based registration frameworks. Additionally, our method models the movement across the entire period of the data, and enables interpolation between the frames of the input modality, making the registration continuous in time.


%
%
%
\bibliographystyle{splncs04}
\bibliography{references.bib}
%




\end{document}